# Semi-supervised estimation of event temporal length for cell event detection

Ha Tran Hong Phan[a], Ashnil Kumar[a], David Feng[a,d], Michael Fulham[b,c], Jinman Kim[a]

[a] School of Information Technologies, The University of Sydney

[b] Sydney Medical School, The University of Sydney

[c] Department of Molecular Imaging, Royal Prince Alfred Hospital

[d] Med-X Research Institute, Shanghai Jiao Tong University

**Abstract**

Cell event detection in cell videos is essential for monitoring of cellular behavior over extended time periods. Deep learning methods have shown great success in the detection of cell events for their ability to capture more discriminative features of cellular processes compared to traditional methods. In particular, convolutional long short-term memory (LSTM) models, which exploits the changes in cell events observable in video sequences, is the state-of-the-art for mitosis detection in cell videos. However, their limitations are the determination of the input sequence length, which is often performed empirically, and the need for a large annotated training dataset which is expensive to prepare. We propose a novel semi-supervised method of optimal length detection for mitosis detection with two key contributions: (i) an unsupervised step for learning the spatial and temporal locations of cells in their normal stage and approximating the distribution of temporal lengths of cell events and, (ii) a step of inferring, from that distribution, an optimal input sequence length and a minimal number of annotated frames for training a LSTM model for each particular video. We evaluated our method in detecting mitosis in densely packed stem cells in a phase-contrast microscopy videos. Our experimental data prove that increasing the input sequence length of LSTM can lead to a decrease in performance. Our results also show that by approximating the optimal input sequence length of the tested video, a model trained with only 18 annotated frames achieved F1-scores of 0.880-0.907, which are 10% higher than those of other published methods with a full set of 110 training annotated frames.

# I/ Introduction

Research in stem cell biology and pharmacology includes the monitoring of cellular behavior over extended time periods. In biomedical experiments, the total count of cell division (mitosis and meiosis) and cell death are used as metrics for studying the health and growth of cell populations [1, 2]. Phase-contrast, time-lapse microscopy (PCM) is the preferred cell imaging modality for capturing such metrics because it is a non-destructive technique that enables in-vitro, long-term and continuous monitoring / analysis of cell populations [3-7], which is in contrast to luminescent, colorimetric or fluorescent assays. PCM data holds tremendous potential for biological and pharmacological discovery but this potential is predicated on the ability to analyse the behavior of the cells within the image data, but at present the analysis process relies on manual human observation and annotation, which is time-consuming, irreproducible, and can be plagued by inter-observer variability. Automated, robust analytical methods for the quantitative analysis of 'cell events' in PCM data are required to improve analysis while reducing time-consuming manual annotations [8, 9]. The state-of-the-art methods in cell event detection trained their models, which learned to determine the locations of the target events, on annotations of cell events.

In this paper, we define as 'cell events' those activities that significantly change the visual properties of a whole cell spatially and temporally, such as shape, size, and movement across the frame. For example, during death a cell's membrane swells into spherical bubbles [10], while during division a cell shrinks into a circular shape [11]. The occurrence of these events are governed by many stochastic biomolecular elements and can vary depending upon the conditions in which the cells grow [12-14]. The visual properties of cells, including cell size, movement speed, duration (temporal length) of a cell event, etc., vary in each new biological condition, such as different cell types, different stimuli at varying doses, or the frame rate at which the PCM data were captured. These variations mean that experts with highly specialized skills are required to manually view and annotate cell events in the imaging data. This leads to expensive costs in data acquisition and annotation, which should be reduced.

One approach to cell event detection was to track individual cells and extract information about cell events from the cell trajectories, which were prone to errors and required careful parameter optimization [8, 15]. Other methods did not generate cell trajectories but searched directly for the location of cell events by modeling the temporal correlations between frames using hand-crafted features, such as the works using Hidden Conditional Random Fields [6, 10, 11, 16-18] and the work by Liu et al. with semi-Markov fields [19]. Methods that are based on deep neural networks (DNNs) have led to an increase in accuracy scores by learning discriminative features directly from in-domain data [20]. Convolutional neural networks (CNN), which extracted visual features from static images [21, 22] and volumetric images [23], were used for mitosis detection, however, they were not designed to model long-range temporal patterns in video data. The leading techniques for time-series data, such as in video analysis and natural language processing, are recurrent neural networks, particularly the long short-term memory (LSTM) method [24-28]. The incorporation of CNN components into LSTM units to form convolutional LSTM (ConvLSTM) units combines visual feature extraction with the capacity to learn dynamic changes over time, thereby providing the state-of-the-art accuracy scores for mitosis detection in PCM videos [29].

There are two problems with using LSTM-based methods in cell event detection:

- The length of the input sequence for LSTM units was determined empirically, without any analytical method [29-31], which may result in sub-optimal accuracy.
- The optimal number of cell video frames needed training an LSTM neural network for cell event detection is unknown and never determined, which may lead to unnecessary expensive data collection and annotation.

The effects of changing the input sequence length for LSTM units have yet to be studied. As the LSTM units need to learn to detect spatio-temporal patterns of cell events, they must see the complete progress of the target cell events as well as that of other cell events with similar spatio-temporal patterns to distinguish between them. If the input sequence is too short, the LSTM units will not learn complete temporal patterns in the progress of the targeted cell events. If the input sequence is too long, in case of limited number of training samples, the LSTM units will be prone to learning irrelevant or redundant patterns of objects or events that are temporally adjacent to the targeted cell events. On the other hand, the optimal input sequence length should cover the temporal lengths of both the true cell events and the potential false positives. In addition, the number of video frames in a training dataset must be equal or higher than the input sequence length for LSTM units. This technical requirement means the input sequence length is the lower bound for the number of frames to be annotated. With the highly changing visual properties in cells, the optimal input sequence length may vary for each cell video with different cell types, drug doses, or video frame rates.

Our aim is to figure out what the durations of cell events are in an unsupervised manner so that the input sequence length for LSTM units and the amount of needed annotated video frames can be estimated. We propose a semi-supervised method for cell event detection that requires minimal amount of training samples to achieve detection accuracy that is equivalent to methods trained with larger datasets. Our method introduces the following technical contributions:

- An unsupervised method for learning the map for detection of cells in normal stage.
- An unsupervised method for learning the transitioning between normal stage and event-occurring stage of cells to predict the existence and approximate the temporal lengths of cell-level events.
- A semi-supervised method that enables learning to map an input image sequence to a spatio-temporal detection of mitotic events, using minimal amount of training data.
- Guidelines for preparing the minimal training dataset and selecting the sequence length for the recurrent neural network for optimal cell event detection.

The framework with three neural network models are shown in Fig. 1.

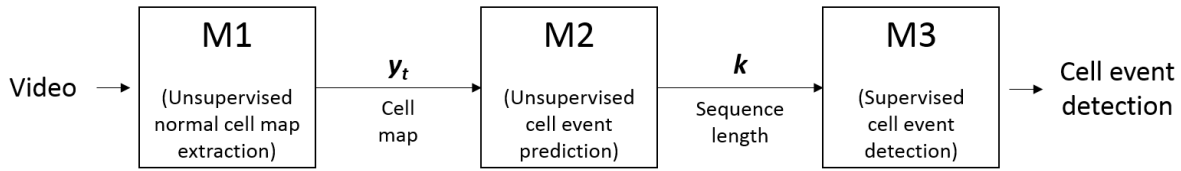

Fig. 1. The framework with three models: M1 for the unsupervised extraction of the map of normal cells $y_t$; M2 for the unsupervised prediction of cell events and estimation of the optimal sequence length $k$ for a recurrent neural network for cell event detection; and M3 for the supervised cell event detection.

## II/ Related works

An approach in cell event detection with PCM focused on methods to track individual cells by matching objects frame-by-frame, and extract information about cell events from the cell trajectories [8, 32-34]. This approach was limited by the variability in cell appearance, the similarities between target and adjacent cells, and the unpredictability of cell movement that resulted in inaccurate detection. Other methods directly targeted the events of interest by learning to classify feature patterns between event and non-event locations in an image sequence. The common methods were based upon Support Vector Machines (SVM) and Hidden Conditional Random Fields (HCRFs) [6, 10, 11, 16-19]. HCRFs proved superior to the SVMs for their ability to model the temporal dynamics of cell events. These methods tended to be application-specific and are often optimized for a particular cell type as they relied on hand-crafted features. Application of these methods required considerable effort to select customized feature sets, when compared with deep learning's capability in learning features from raw data [35]. A candidate selection step was always implemented to reduce the search space, which relied on prior knowledge of the particular imaging data, making SVMs and HCRFs methods less generalizable than their deep learning counterparts [29].

Deep learning approaches have been used for mitosis detection in static images, such as the 2D CNN for breast cancer histology images in Ciresan et al. [36] and another 2D CNN for static images in

Shkolyar et al. [21]. For mitosis detection in a PCM sequence, Nie et al. [23] reported a 3D CNN for volumetric regions, and Mao et al. [22] used hierarchical CNN that analyzed both original and motion images. Combinations of separate CNN and LSTM components allow for modeling of spatio-temporal patterns, as reported in Mao et al. [30] and Su et al. [31]. Joint feature learning and temporal modeling was also applied in the probabilistic model reported by Liu et al. [37]. Recent advances in deep learning have made capturing of spatio-temporal patterns feasible by extending the fully-connected LSTM to have convolutional structures, as in ConvLSTMs [38-40]. Unsupervised and supervised methods developed upon ConvLSTM structure, with end-to-end training property, have achieved state-of-the-art accuracy scores in mitosis detection in Phan et al. [29]. These methods required empirical determination of the input sequence length for LSTM units.

The abovementioned supervised methods always rely on a large number of training samples to gain optimal accuracy. Other methods that aimed to require less training data have also been reported. A multioutput Guassian process model was used to detect generic event after training with a small set of event-free training data [41]. Preparing a training time-lapse PCM free of mitotic events is challenging as controlling this type of stochastic cell events without changing a cell population's characteristics might prove impossible. Kandemir et al. [42] reported another multioutput Gaussian process method with active-learning that need only 3% annotated data. This method, however, required experts to be actively involved in the training process to help the model classify proposed candidates. These approaches relied on hand-crafted features that were proven to be inferior to the learned features by deep learning [20]. In addition, they also were not able to specify, with an analytical method, how much training data to prepare, which may assist domain experts in planning for their experiments and annotation.

## III/ Methods

A/ Conceptual formulation

Let $x_t \in [0, 1]^{M \times M}$ represent a microscopy video frame of size $M \times M$ that contains a cell population at time $t$. The problem of approximating the temporal lengths of cell-level events in an unsupervised manner can be formulated in two steps: (i) learning to extract a map that represents the presence of cells in their normal stage, and (ii) learning to predict the occurrences of cell-level events by speculating from the extracted map. As PCM images do not provide observable information about the structure or biomolecular factors inside cells, the model is constrained to relying on the observable visual properties rather than the internal cell biological properties. Our aim is to solve the above two steps without any ground truth to train models. A solution for the step (i) is to perform a spatial reconstruction task to find $P(y_t|x_t)$, where $y_t \in [0, 1]^{M \times M}$ is the map that represents the presence of cells in their normal stage. After having obtained $y_t$, a solution for the step (ii) is to perform a spatio-temporal reconstruction task to find $P(y_{k:t}|y_{1:k})$, where $x_{k:t}$ are the frames containing cell events and $x_{1:k}$ contain the frames before the occurrences of the cell events, respectively.

The map $y_t$ must represent biologically meaningful structures that are associated with individual cells. $y_t$ can be extracted from the hidden state learned in a spatial reconstruction task that finds $P(h_t|x_t)$ and $P(x'_t|h_t)$, where $x'_t$ is the reconstructed frame that is intended to be closely similar to $x_t$, in an unsupervised manner similar to autoencoder [43]. $h_t \in [0, 1]^{n \times M \times M}$, with $n$ being the number of feature maps, is the hidden state that represents the spatial structure of the imaging data. At each pixel in the frame of size $M \times M$, there should be only one out of $n$ feature maps activated, while the rest of them are given zero values. For the feature maps of $h_t$ to represent biologically meaningful structures, it should be possible to reconstruct each one of them given the other feature maps, as follows:

$$P(h_t^n|x_t) = P(h_t^n|h_t^{n-1},\dots,h_t^1) \prod_{k=1}^{n-1} P(h_t^k|x_t). \qquad (1)$$

With this constraint, $y_t$ can be extracted as one of the feature maps of $h_t$. To fulfill this requirement, a feature map in $h_t$ can be dropped out in a random manner when performing the reconstruction task, encouraging the model to learn structurally related representations.

Being an extraction from an image sequence, $y_t$ contains temporal patterns that are associated with cell events. A spatio-temporal reconstruction task can be performed to find $P(u_t|y_{1:t})$ and $P(y'_t|u_t)$, $y'_t$ is the reconstructed frame that is intended to be closely similar to $y_t$, and $u_t$ is the hidden state that represents the frames $y_{1:t}$. Markov processes are a well-established technique to model temporal activities in cell populations. For example, a Markov model was used for stochastic state transitions of cancer cells in Gupta et al. [44], or a spatial Markov process with probabilities of movements for individual cells in Altrock et al. [45]. The hidden state $u_t$ can be modelled as a Markov random field $u_t \in [0,1]^{o \times M \times M}$, with $o$ being the number of feature maps. Once a cell event happens, the representation of the cell in $y_{1:k}$ disappears in $y_{k:t}$. Two components can be derived from $u_t$. A component $e_t \in [0,1]^{p \times M \times M}$, with $p$ being the number of feature maps, is the hidden state for cells undergoing cell events that disappear in $y_{k:t}$. Another component $c_t \in [0,1]^{q \times M \times M}$, with $q$ being the number of feature maps, is the hidden state containing context information about normal cells, including shapes, locations, and movements, that can be learned from $y_{1:k}$. $e_t$ and $c_t$ contribute to the formation of $y_t$, as shown in Fig. 2.

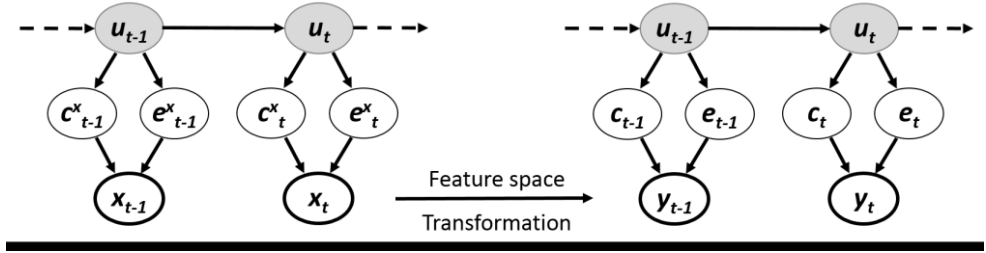

Fig. 2. Our method transforms the original frame $x_t$ into the map of normal cells $y_t$, thus turning the hidden states representing the context information $c_t^x$ and cell events $e_t^x$ of $x_t$ into the equivalent context information $c_t$ and cell events $e_t$ of $y_t$. Both $x_t$ and $y_t$ are the expressions of the hidden state from the Markov random field $u_t$ that governs the activities of the cell population.

B/ Mathematical formulation

The state of the cell population can be captured by the joint probability distribution

$$P(y_{1:N}, e_{1:N}, c_{1:N}, u_{1:N}) = \prod_{t=1}^{N} P(y_t|e_t, c_t) P(e_t|u_t) P(c_t|u_t) P(u_t|u_{t-1}) \qquad (2)$$

where

- $P(u_t|u_{t-1})$ denotes the hidden state transition probability capturing the state of the cell population;
- $P(e_t|u_t)$ models the relationship between the occurrences of cell events and the state of the cell population;
- $P(c_t|u_t)$ models the relationship between the context information and the state of the cell population;
- $P(y_t|e_t, c_t)$ models the formation of the map of normal cells in the cell population.

We can apply recursive Bayesian estimation [46] to obtain the belief $B_t^u$ which, at any point in time $t$, corresponds to a distribution $Bel(u_t) = P(u_t|x_{1:t})$.

$$Bel^-(u_t) = \int_{u_{t-1}} P(u_t|u_{t-1}) Bel(u_{t-1}) \qquad (3)$$

$$Bel(u_t) \propto \int_{e_t} \int_{c_t} P(y_t|e_t, c_t) P(e_t|u_t) P(c_t|u_t) Bel^-(u_t) \qquad (4)$$

where $Bel^-(u_t)$ denotes the predicted belief one time step forward after the map $y_{t-1}$ at time $t-1$, and $Bel(u_t)$ denotes the corrected belief based on the prediction $Bel^-(u_t)$ in light of the new map $y_t$ at time $t$. The two phases, which are expressed by (3) and (4), can be called the "predict" and "update" steps/functions that alternate to gain an accurate knowledge about the map of normal cells.

$P(e_t|y_{1:t})$ and $P(c_t|y_{1:t})$ can then be expressed as:

$$P(e_t|y_{1:t}) = \int_{u_t} P(e_t|u_t) Bel(u_t) \qquad (5)$$

$$P(c_t|y_{1:t}) = \int_{u_t} P(c_t|u_t) Bel(u_t) \qquad (6)$$

In a similar manner to Ondruska et al. [39], we can iteratively calculate the corrected belief $B_t^u$ at current time $t$ with a function $F$:

$$B_t^u = F(B_{t-1}^u, y_t) \qquad (7)$$

defined by (3), (4).

$P(e_t|y_{1:t})$ and $P(c_t|y_{1:t})$ can be calculated using the corrected belief $B_t^u$ which is achieved using (7):

$$P(e_t|y_{1:t}) = P(e_t|B_t^u) \qquad (8)$$

$$P(c_t|y_{1:t}) = P(c_t|B_t^u) \qquad (9)$$

The spatio-temporal reconstruction task can then be modelled with $P(e_t|y_{1:t})$, $P(c_t|y_{1:t})$, and $P(y'_t|e_t, c_t)$, where $y'_t$ is the reconstructed frame that is intended to be closely similar to $y_t$.

Neural networks can be used to model $P(h_t|x_t)$, $P(x'_t|h_t)$, $F(B_{t-1}^u, y_t)$, $P(e_t|B_t^u)$, $P(c_t|B_t^u)$, and $P(y'_t|e_t, c_t)$. For the first model for extracting $y_t$, $P(h_t|x_t)$ and $P(x'_t|h_t)$ can be modelled with CNNs, max-pooling, and a layer for dropping one out of $n$ feature maps. For the second model that learn $e_t$, we assume $B_t^u$ can be represented as vectors, $B_t^u \in \mathbb{R}^{r \times M \times M}$, with $r$ being the number of feature maps. Then $F(B_{t-1}^u, y_t)$ can be a recurrent neural network with $B_t^u$ being the network's memory that get updated after each time step [47]. Feed-forward neural networks can be used to approximate $P(e_t|B_t^u)$, $P(c_t|B_t^u)$, and $P(y'_t|e_t, c_t)$. In each of the two models, the neural networks can be combined together and trained as a single neural network in an end-to-end manner.

For the first model, the learning process minimizes a cross-entropy loss of the reconstructed frame $x'_t$ and the original frame $x_t$.

$$L(x_t, x'_t) = -\sum_{i,j} x(i,j)_t \log x'(i,j)_t + (1 - x(i,j)_t) \log(1 - x'(i,j)_t) \qquad (10)$$

where $i$ and $j$ are the row and column indices of a pixel $x(i,j)_t$ in the frame $x_t$. $x'_t$ can be derived using two functions: an encoder function $h_t = f(x_{1:t-1}, x_{t:t+k})$ that models $P(h_t|x_t)$, and a decoder function $x'_t = g(h_t)$ that models $P(x'_t|h_t)$ and produces the reconstructed frame $x'_t$.

Similarly for the second model, the learning process minimizes a cross-entropy loss of the reconstructed map $y'_t$ and the original map $y_t$.

$$L(y_t, y'_t) = -\sum_{i,j} y(i,j)_t \log y'(i,j)_t + (1 - y(i,j)_t) \log(1 - y'(i,j)_t) \quad (11)$$

where $i$ and $j$ are the row and column indices of a pixel $y(i,j)_t$ in the frame $y_t$. $y'_t$ can be derived using two functions: an encoder function $(e_t, c_t) = f(y_{1:t})$ that models $F(B^u_{t-1}, y_t)$, $P(e_t|B^u_t)$, and $P(c_t|B^u_t)$, and a decoder function $y'_t = g(e_t, c_t)$ that models $P(y'_t|e_t, c_t)$ and produces the reconstructed map $y'_t$.

In practice, for a sequence of $t$ frames, we use the loss function:

$$L(y_{1:t}, y'_{1:t}) = -\sum_{1}^{t}\sum_{i,j} y(i,j)_t \log y'(i,j)_t + (1 - y(i,j)_t) \log(1 - y'(i,j)_t) \quad (12)$$

where the encoder function $f(y_{1:t})$ outputs $(e_{1:t}, c_{1:t})$, and the decoder function $g(e_{1:t}, c_{1:t})$ reconstructs $y'_{1:t}$. By reconstructing a subsequence $y'_{1:t}$ instead of a single frame $y'_t$, more information about the error is back-propagated to guide the updating of the neural network's parameters.

As the training process is unsupervised, no ground truth is available for learning the hidden state of cells undergoing events $e_t$. This problem can be solved by artificially generating the ground truth for learning $e_t$ and $c_t$. 3D regions that are associated with trajectories of separate individual cells in the map of normal cells $y_t$ can be formed by matching temporally overlapping regions of 2D frames of the map together. Random parts of these 3D regions can be removed to simulate occurrences of cell events, where cells disappear from the map of normal cells $y_t$ and the same cells or their daughter cells (mitosis) re-appear after the cell events, as explained in Fig. 3. These removed parts can be assumed as the annotations for the cell events $e_t$, and the remaining parts of the map $y_t$ can be considered as the context information $c_t$. The decoder function can then be simplified as a summation of $e_t$ and $c_t$ to produce $y'_t$.

Neural networks can be used to extract these hidden states from an image sequence. For the first model for extracting $y_t$, $P(h_t|x_t)$ and $P(x'_t|h_t)$ can be modelled with CNNs, max-pooling, and a layer for dropping one out of $n$ feature maps. $y_t$ can then be extracted as one of the feature maps of $h_t$ For the second model that learn $e_{1:t}$, a recurrent neural network can be combined with CNNs to extract spatial and temporal information from the maps of normal cells $y_{1:t}$, and use the artificially generated ground truth context information $c_{1:t}$, as explained above, to learn to predict cell events $e_{1:t}$. In each of the two models, the neural networks can be combined together and trained as a single neural network in an end-to-end manner.

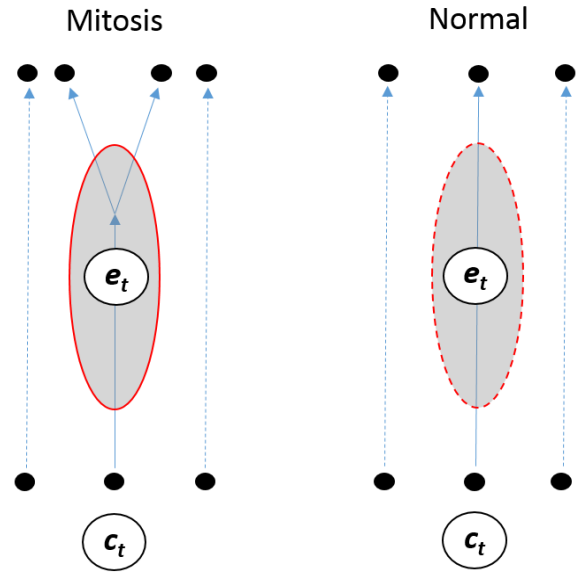

Fig. 3. Illustration of the artificial generation of training data that simulates the occurrences of cell events. The contextual hidden state $c_t$ for normal cells (blue arrows) includes neighbor cells and the event undergoing cells before and after the occurrence of a cell event. The hidden state $e_t$ for cells undergoing events (red circles) contains only the cell events (left side), or the artificial removal of parts of the cell trajectories to emulate cell events (right side).

C/ Model for extraction of normal cell map

Fig. 4 shows our M1 model for the extraction of the normal cell map $y_t$. The encoding part consists of three CNN layers, a softmax and max-pooling layer, and a dropout layer that deactivate one random map of the $n$ feature maps of $h_t$. The feature maps extracted are then added with a noise of a uniform distribution in the range [-0.2, 0.2] to encourage the neural network to only learn essential biological structures of cells instead of memorizing the input frame $x_t$. The decoding part, with three CNN layers, then reconstructs the original frame to generate $x'_t$ from the dropped-out feature maps of $h_t$. As there is no ground truth to learn the map of normal cells $y_t$, $y_t$ can be selected from the $n$ feature maps of $h_t$ based on knowledge of the internal part of cells as disconnected and round shaped objects. A binarization step, which changes all non-zero values into ones, and an erosion step are then applied to the map $y_t$ for noise removal. The CNN layers have kernel sizes of 5×5, except the last CNN layer with a kernel size of 1×1.

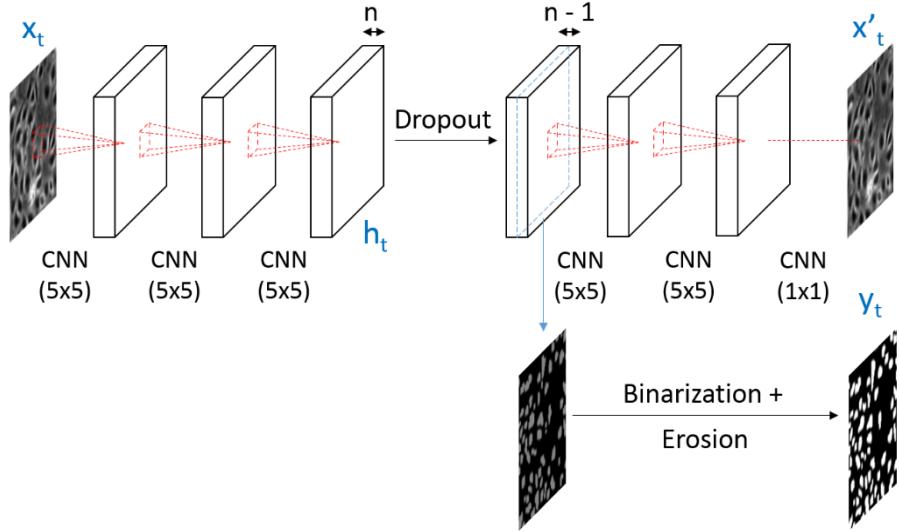

Fig. 4. Unsupervised model M1 for the extraction of the map of normal cells $y_t$. The model takes as input the original frames $x_t$ and outputs the reconstructed frame $x'_t$ and the feature maps of $h_t$. $y_t$ is then selected as one of the $n$ feature classes of $h_t$ and processed with binarization and erosion steps.

D/ Model for cell event occurrence prediction

The original LSTM unit [26] used full connections in input-to-state and state-to-state transitions that do not preserve spatial information in image sequences. For learning spatio-temporal patterns in cell events and context information, the ConvLSTM, reported in Xingjian et al. [38], is chosen as our baseline unit due to its preservation of both spatial and temporal information in modelling long-range dependences. Fig. 5 shows our M2 model for the prediction of the occurrences of cell events, which comprises of: two layers of CNN for spatial feature extraction from the map of normal cells $y_{1:t}$; a layer of ConvLSTM for learning the context information, including the locations, shapes, and movements of objects in $y_{1:t}$, and predicting the occurrences of cell events in its outputted hidden states; and a layer of CNN for decoding the hidden states from ConvLSTM to generate the prediction of cells undergoing events $e'_{1:t}$. During training, the input sequence is the context sequence $c_{1:t}$, and the label sequence is the event sequence $e_{1:t}$ that are artificially generated by removing random parts of the 3D regions in $y_{1:t}$, which are associated with trajectories of individual cells. During inference, the original sequence $y_{1:t}$ is used as the input sequence so that the model can predict the actual cell events based on the context information provided in $y_{1:t}$. The ConvLSTM units have 16 hidden states. Both ConvLSTM units and CNN layers have kernel sizes of 3×3, except the last CNN layer that has a kernel size of 1×1.

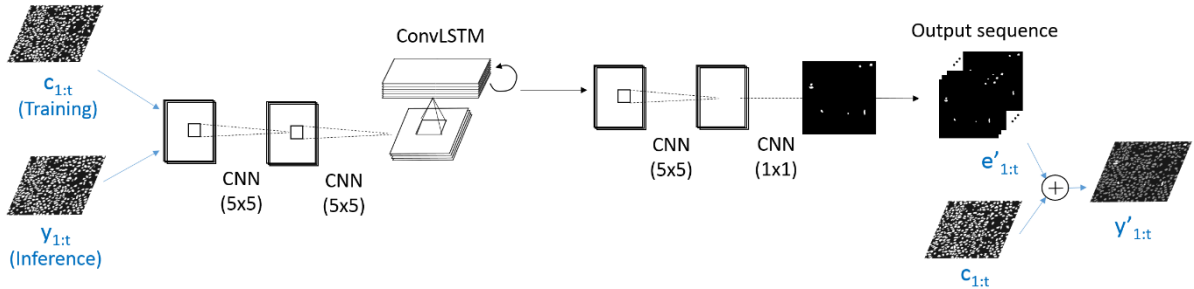

Fig. 5. Unsupervised model M2 for learning the prediction of cell event occurrences $e'_{1:t}$. Two CNN layers for initial feature extraction from input frames, being the hidden state for normal cells'

locations $y_{1:t}$ during inference, and the artificially generated hidden states for contextual information of normal cells $c_{1:t}$ during training. A ConvLSTM unit learns spatio-temporal patterns of cell events and context information from the outputs of these CNN. Two CNN layers decode the learned patterns into a map of cell event occurrences $e'_{1:t}$.

E/ Post-processing step for cell event occurrence prediction

The prediction of cell events $e'_{1:t}$ is processed to estimate the temporal lengths of sequences of cell events. Binarization is applied for $e'_{1:t}$, with values above 0.5 set to 1.0 and the rest set to 0.0. Temporally overlapping areas in the frames of the sequence $e'_{1:t}$ are combined together to form 3D regions of the cell events. The temporal lengths of these 3D regions are then used to determine the appropriate sequence length $k$ for the ConvLSTM units in the subsequent supervised training step for M3. The statistics of these temporal lengths of predicted cell events, such as mean, 50 percentile, and 75 percentile, are calculated to provide the overall information about the cell population.

F/ Supervised model for mitosis detection

After having known the overall properties of predicted cell events, particularly their temporal lengths, a model based on recurrent neural networks can be developed with a goal of minimizing training data. Our supervised neural network, M3, has two main parts as shown in Fig. 6. The encoding part consists of two ConvLSTM units: one reading the raw image sequence input forward in time; the other reading it in reverse. As mitotic events happen in phases, the complementary temporal information gained in this two-way extraction technique leads to better temporal localization of mitotic phases. The hidden states outputted by these two ConvLSTM units are concatenated together. The extracted features are then passed through a dropout layer, with a dropout rate of 0.3, to reduce overfitting [48] in case of limited training dataset. The output from this layer contains spatial and temporal information for the subsequent CNN layers, which form the decoding part. These CNN layers generate the 3D mitosis detection output. The ConvLSTM units have 16 hidden states. Both ConvLSTM units and CNN layers have kernel sizes of 3×3, except the last CNN layer that has a kernel size of 1×1.

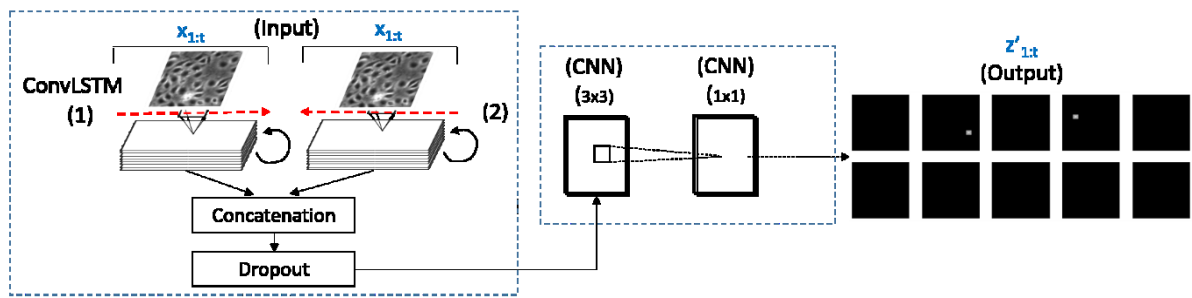

Fig. 6. Supervised model M3 for mitosis detection. The model comprises two ConvLSTM units with one unit reading the input sequence in forwards direction and another unit reading in reverse direction, a dropout layer to avoid overfitting, and two CNN layers for generating the 3D map of mitosis detection.

G/ Training data preparation and output post-processing for supervised mitosis detection

The training dataset is an image sequence that consists of subsequent frames from the original cell video. This sequence must have the number of frames being equal or higher than the sequence length

$k$ of the ConvLSTM units. The training frames are then annotated for mitosis to get the ground truth for the supervised training. The annotated pixels, one for each mitotic event, are temporally and spatially expanded with binary dilation three times to form a volume around the point of mitosis, where two daughter cells are formed and start to separate.

The model was trained by minimizing the cross-entropy loss between the model's output and the labels of mitosis. The loss function is as follows:

$$L(z_{1:t}, z'_{1:t}) = -\sum_{1}^{t}\sum_{i,j} z(i,j)_t \log z'(i,j)_t + (1 - z(i,j)_t) \log(1 - z'(i,j)_t) \quad (13)$$

where $z_{1:t}$ is the label sequence and $z'_{1:t}$ is the model's output for the input frames $x_{1:t}$ from time 1 to time $t$; and $i$ and $j$ are the row and column indices of a pixel $z(i,j)_t$ in the label frame $z_t$. Binarization is applied for $z'_{1:t}$, with values above 0.5 set to 1.0 and the rest set to 0.0. The output of the model, $z'_{1:t}$, indicates the spatial and temporal areas where mitosis occurs. During inference, the input sequences are generated using a temporal step of 2 frames, and the outputs are summed together according to the spatial and temporal coordinates of the input sequences. A 3D map of mitosis detection is then generated after a binarization step for the summed output for the whole testing sequence, with non-zero values indicating the occurrence of mitosis. The exact pixel associated with the center of each detected mitotic event is determined as the mass center of each group of spatially and temporally adjacent detected pixels.

H/ Implementation

Our method was implemented using the Tensorflow framework and an NVIDIA 12GB Titan X (Maxwell architecture). The video frames were pre-processed by linearly rescaling image pixel values to be between 0 and 1. All of the input frames $x_{1:t}$ and maps $y_{1:t}$ are downscaled by a factor of 4 from the original size 1040×1392 pixels of the video frames, into the size of 260×348. This reduces the large memory footprint required for processing the original frames. The output of M3, with size 260×348, is rescaled back to the original size 1040×1392 before the exact pixels of mitotic events are determined. We used a data augmentation method to reduce the likelihood of overfitting during training [49]. For the training dataset for each model, we augmented the samples by flipping each sequence horizontally and vertically, and by rotation by 90, 180, and 270 degrees. Using entire frames to form the input sequences can retain the context information of surrounding cells that can be important for the model to distinguish mitosis from similarly appearing events. The models take as input the entire sequences of consecutive complete frames. For CNN and ConvLSTM layers, zero-padding was performed on the hidden states to assume no prior knowledge about the cells outside of the boundary of the video frame. All models were trained using RMSProp [50] and back-propagation through time [43]. The learning rate was 10-3 and a decay rate was 0.9. Weights were initialized using Xaivier initialization [51] and biases were initially set to be zero. The training iteration number for model M3 is 6000 for all experiments.

I/ Dataset

We used a video of 210 frames of densely packed stem cell PCM images, which is considered more difficult for mitosis detection than sparsely located cell images [20]. The video captured bovine aortic endothelial cells (BAEC) that was also used by Huh et al. [11]. The video starts with regions of densely packed cells on the sides and an empty area in the middle, and after cell divisions, the video ends with cells completely covering the frame. True positive detection of mitosis in our paper is consistent with Huh et al. [11], where a point of the complete separation of two daughter cells is correctly detected if it is spatially within 10 pixels and temporally within one or three frames from the manual annotation.

J/ Experiments

We trained our models, M1 for the extraction of the normal cell map $y_t$, M2 for the prediction of cell event occurrences, and M3 for supervised mitosis detection, on the first 110 frames of the video and then tested the detection accuracy of M3 on the remaining last 100 frames. The number $n$ of feature maps of $h_t$ was set to be 6. The statistics of the temporal lengths of the predicted cell events were calculated in the post-processing step for the output of M2. The ConvLSTM units was used to learn to distinguish among the temporal patterns associated with phases of cell events, and the temporal patterns of activities similar to, but not being, the target cell events. Therefore, the sequence length $k$ of the ConvLSTM units should be an upper bound of a range that covers the majority of cell event temporal lengths in the video. With the mean and 75 percentile of predicted cell events being 6.38 and 8.0, respectively (Table I), the value of $k$ should be about 8.

We conducted experiments with $k$ being assigned values in the range of 4, 6, 8, 10, 14, and 18. We intend to minimize the number of frames to be annotated for training M3, which must be at least the sequence length $k$. The training frames for M3 were taken from the last frames from the list of the first 110 frames of the video and the evaluation was performed on the remaining last 100 frames. We used this standard evaluation approach as used by Huh et al. [11], which used the first 110 frames from a video of 210 frames for training and the last 100 frames for testing in their fully supervised method, to allow a comparison to prior work using the same dataset. The number of frames were varied in the range of 4, 6, 8, 10, 14, 18, 22, and 110. Mitosis detection by M3 was evaluated in terms of F1-score, which is a balanced measure for precision and recall, for two allowed timing errors being one or three frames.

## IV/ Results

A/ Extraction of normal cell map

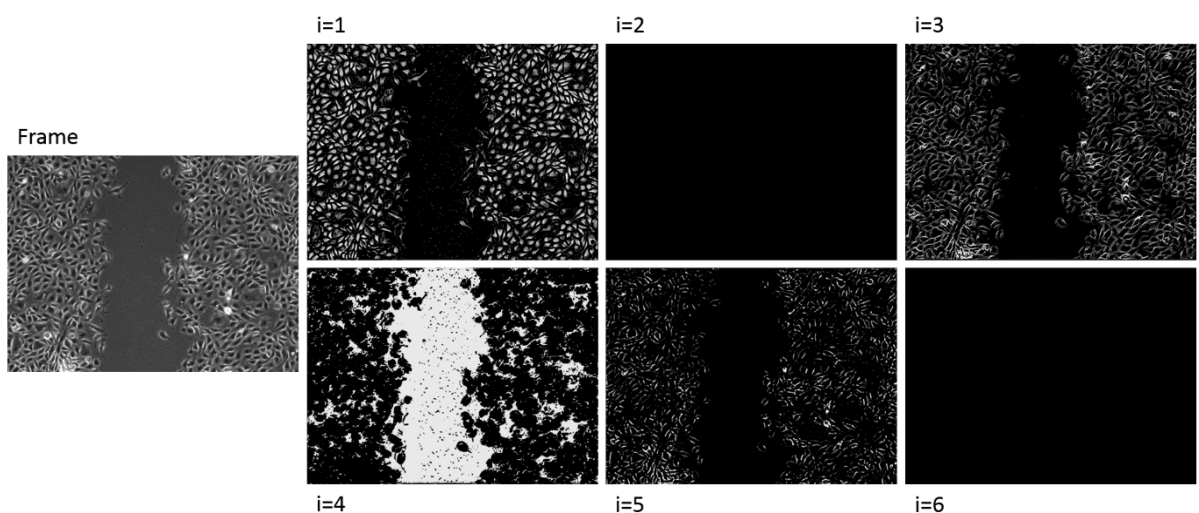

Fig. 7. An example of the original frame $x_t$ (left side) and its six feature maps of $h_t$ (right side). Four maps contain four separated biologically meaningful structures, and two maps are blank. This phenomenon is discussed in the section V.

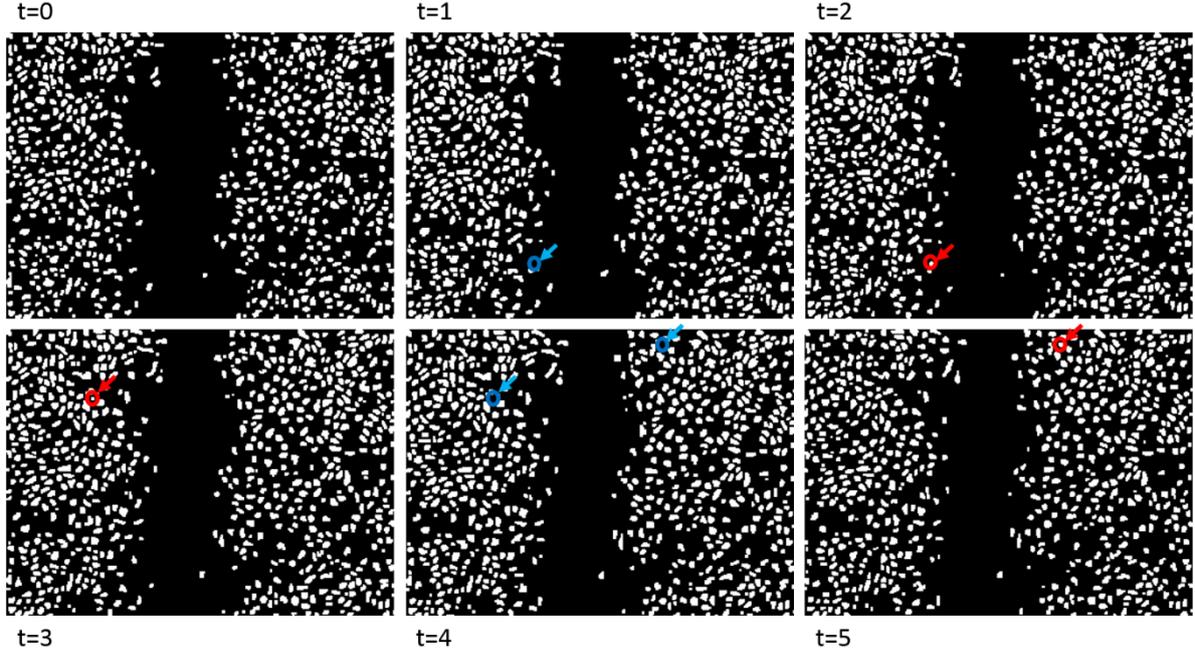

Fig. 8. An example of a sequence of the maps $y_t$ during six time steps. Blue circles and arrows indicate the locations of cells that are not in the map, and red circles and arrows indicate the locations of the same cells that reappear or about to disappear from the map.

Fig. 7 shows an example of the original frame $x_t$ on the left side and its six feature maps of $h_t$ on the right side. The six feature maps show separated biologically meaningful structures of the cell population, and one of them is selected to undergo binarization and erosion steps to form the map of normal cells $y_t$. Among the six feature maps, four were activated and two remained blank. This is intended because the number of feature maps was overestimated to ensure all structures were captured and separated. An explanation for this can be found in the section V. Fig. 8 shows a randomly selected sequence of the maps $y_t$ during six time steps.

B/ Cell event occurrence prediction

Table I: Statistics of the temporal lengths of all predicted cell events.

|  | Average | Standard deviation | 50 percentile | 75 percentile |
|---|---|---|---|---|
| Temporal lengths of all predicted events | 6.38 | 5.63 | 5.0 | 8.0 |

Table I shows the statistics of the temporal lengths of the predicted cell events $e'_t$ in the first 110 frames of the video.

C/ Supervised mitosis detection

Table II: Comparison of the F1-scores of mitosis detection among M3 models with different values of the sequence length $k$ and different numbers of training frames and the total number of annotated mitotic events within these frames. Results are given across standard annotation thresholds ($th = 1$ and $th = 3$); the highest scores for each set of training frames are highlighted in red.

| | th=1 | | | | | | | |
|---|---|---|---|---|---|---|---|---|
| | Frames=4 Events=7 | Frames=6 Events=15 | Frames=8 Events=20 | Frames=10 Events=23 | Frames=14 Events=32 | Frames=18 Events=42 | Frames=22 Events=52 | Frames=110 Events=191 |
| $k = 4$ | 0.802 | 0.790 | 0.789 | 0.788 | 0.772 | 0.806 | 0.821 | 0.750 |
| $k = 6$ | - | 0.824 | 0.835 | 0.819 | 0.843 | 0.859 | 0.863 | 0.840 |
| $k = 8$ | - | - | 0.824 | 0.836 | 0.867 | 0.880 | 0.870 | 0.885 |
| $k = 10$ | - | - | - | 0.800 | 0.860 | 0.873 | 0.871 | 0.883 |
| $k = 14$ | - | - | - | - | 0.790 | 0.850 | 0.864 | 0.884 |
| $k = 18$ | - | - | - | - | - | 0.816 | 0.870 | 0.908 |
| $k = 22$ | - | - | - | - | - | - | 0.814 | 0.878 |
| | th=3 | | | | | | | |
| | Frames=4 Events=7 | Frames=6 Events=15 | Frames=8 Events=20 | Frames=10 Events=23 | Frames=14 Events=32 | Frames=18 Events=42 | Frames=22 Events=52 | Frames=110 Events=191 |
| $k = 4$ | 0.837 | 0.852 | 0.824 | 0.812 | 0.785 | 0.830 | 0.841 | 0.776 |
| $k = 6$ | - | 0.841 | 0.853 | 0.854 | 0.867 | 0.879 | 0.885 | 0.876 |
| $k = 8$ | - | - | 0.847 | 0.877 | 0.895 | 0.907 | 0.898 | 0.893 |
| $k = 10$ | - | - | - | 0.839 | 0.890 | 0.902 | 0.896 | 0.903 |
| $k = 14$ | - | - | - | - | 0.851 | 0.888 | 0.902 | 0.911 |
| $k = 18$ | - | - | - | - | - | 0.849 | 0.898 | 0.917 |
| $k = 22$ | - | - | - | - | - | - | 0.882 | 0.916 |

Table III: Comparison of the F1-scores of mitosis detection for our method to other methods, across standard annotation thresholds ($th = 1$ and $th = 3$); the best results are in red and second best results in blue.

| | $th = 1$ | | | | | | | | |
|---|---|---|---|---|---|---|---|---|---|
| | M3 ($k = 18$, Frames=110) | M3 ($k = 8$, Frames=18) | M3 ($k = 8$, Frames=14) | M3 ($k = 8$, Frames=10) | Supervised ConvLSTM [29] ($k = 10$, Frames=110) | Unsupervised ConvLSTM [29] ($k = 10$, Frames=110) | HCRF [11] (Frames=110) | EDCRF [11] (Frames=110) | EDCRF [11] (Frames=110) |
| Precision | 0.959 | 0.924 | 0.918 | 0.852 | 0.908 | 0.767 | 0.622 | 0.641 | 0.796 |
| Recall | 0.861 | 0.840 | 0.821 | 0.821 | 0.708 | 0.578 | 0.604 | 0.654 | 0.650 |
| F1-score | 0.908 | 0.880 | 0.867 | 0.836 | 0.795 | 0.659 | 0.613 | 0.647 | 0.716 |
| | $th = 3$ | | | | | | | | |
| | M3 ($k = 18$, Frames=110) | M3 ($k = 8$, Frames=18) | M3 ($k = 8$, Frames=14) | M3 ($k = 8$, Frames=10) | Supervised ConvLSTM [29] ($k = 10$, Frames=110) | Unsupervised ConvLSTM [29] ($k = 10$, Frames=110) | HCRF [11] (Frames=110) | EDCRF [11] (Frames=110) | EDCRF [11] (Frames=110) |
| Precision | 0.971 | 0.963 | 0.955 | 0.894 | 0.913 | 0.856 | 0.695 | 0.718 | 0.847 |
| Recall | 0.869 | 0.858 | 0.843 | 0.861 | 0.712 | 0.644 | 0.675 | 0.733 | 0.692 |
| F1-score | 0.917 | 0.907 | 0.895 | 0.877 | 0.800 | 0.735 | 0.685 | 0.726 | 0.761 |

In Table II, we show the highest F1-scores achieved by training M3 with varying values of the sequence length $k$ for the ConvLSTM units and number of frames and annotated mitotic events in the training dataset. Only experiments with the number of frames equal or larger than the sequence length $k$ were conducted. Table III shows the comparisons between our method and other published methods for mitosis detection in densely packed stem cell PCM images.

## V/ Discussion

The main findings in our experiments were: (i) our unsupervised model M1 was able to extract maps of biologically meaningful structures, including a map of locations of cells in their normal stage; (ii) our unsupervised model M2 predicted cell events and estimated the optimal temporal length for designing recurrent neural networks for event detection; and (iii) by leveraging the estimated optimal temporal length, minimal annotations for training our supervised model M3 were prepared to gain a mitosis detection accuracy that approaches the accuracy of training with a fully annotated dataset.

The feature maps extracted by M1 shows that the model has learned to separate pixels into groups of biologically meaningful structures. The evidence is in the visually consistent structures captured in four of the six maps, with the map ($i = 1$) corresponding mostly to the locations of the part of normal cells, which is the cytoplasm and nucleus inside the cell membrane, the map ($i = 3$) and ($i = 5$) showing mainly the edges of cells, and the map ($i = 4$) showing the background (Fig. 7). The other two maps ($i = 2$) and ($i = 6$) were not activated. The fact that M1 used only four maps, instead of all the allowed maps, demonstrates that it was not trying to copy the input frame in ways such as mere clustering based on pixel intensity, and was learning spatially correlated structures, as explained with the equation (1). It is a good practice to heuristically run experiments with varying numbers of allowed maps on the dataset and overestimate the number to be higher than the number of activated maps to ensure that biologically meaningful structures are completely separated from each other. With the same number of six maps, a cell population with more complex structures would activate all the maps. Fig. 8 shows the well-separated objects in the map $y_t$, which are associated with the locations of normal cells, and the capture of their temporal dynamics, which were the disappearances and returns of objects highlighted with arrows and circles.

The model M2 learned the spatio-temporal patterns associated with cell events on an artificially generated training dataset, and predicted cell event occurrences on a real set of frames. The training dataset was created by taking out parts of the trajectories of cells in the map $y_t$. All predicted events were of importance as they can be instances that are similar to mitotic events and give good examples for the model to learn to distinguish false positives from true positives. A good training dataset should include both mitotic and non-mitotic cell events. We hypothesize that the statistics of all the predicted cell events, such as the mean and 75 percentile, give a good reference for designing the supervised experiments. This hypothesis has been proven with the experimental results of the supervised mitosis detection by M3. In our experiments, the 75th percentile value of sequence length $k$=8, was found be the best for the recurrent neural network. In Table II, the F1-scores are often highest when all 110 frames were used for training, except for the cases of $k = 4$ and $k = 6$ which are below the mean value 6.38 in Table I, proving the usefulness of the mean value as a lower bound for the sequence length $k$. For the training datasets of 10, 14, and 18 frames, the model with an input sequence length of 8 for the ConvLSTM units always gain the highest F1-scores, confirming that number 8 is the optimal value, confirming the 75th percentile as a good reference for the sequence length $k$. The highest possible F1-scores were achieved with $k = 18$ and all 110 frames for both allowed temporal errors of one or three frames, which is expected for a full set of annotations with excessively sufficient examples for training a ConvLSTM unit with a long sequence length for learning all distinguishing spatio-temporal features of mitosis. Annotating a whole dataset, however, is expensive and not practical as discussed above. With a dataset of 18 frames, the model with $k = 8$ gives F1-scores only about 1-3% lower than the highest possible F1-scores after training with 18 frames, and about 3-4% lower after training with 14 frames. A noticeable phenomenon is that F1-score often drops to about 0.85 or below when the number of frames is equal to the sequence length $k$. This is possibly due to the lower variance in the error signals the model received when the same input sequences are used repeatedly. The results in Table II shows that ever increasing the number of training frames or mitotic annotations does not give a proportional increase in detection accuracy. Our experimental observations produced the following guidelines for designing a supervised method with RNN units:

- The optimal sequence length should be on the high end of the percentile ranks of the distribution of the temporal lengths of cell events. For example, a percentile rank should be ≥ 70, to cover all the major variations in the temporal patterns in the analyzed sequence. An optimal sequence length for RNN units should also be a value that is high enough to include all the temporal patterns in the input sequence, for both mitotic events and those with similar visual features.
- With limited training samples, a sequence length that is longer than the optimal value can result in a lower accuracy. Having analyzed data in an excessively long sequence might lead to overfitting by the model giving high probability to spatio-temporal patterns of non-event instances that occurs temporally adjacent to the actual cell events. This can be observed in the changes in F1-scores in Table II when different values of k are used.
- To minimize data annotation while gaining an optimal detection accuracy, it is recommended that the sequence length for the recurrent neural network is set to be the 75 percentile of the temporal lengths of predicted cell events, and the number of frames to be annotated is a few frames more than the sequence length. This is the case of k=8 with 14 training frames.

It has well recognized that higher accuracy in cell event detection can be achieved with more training data; however, as we have previously noted, the requisite scale of annotated training data is rarely available. Our study has demonstrated a more economical approach that produces comparable outcomes without the requirement for extensive training data. Without knowing the distribution of the temporal lengths of cell events or obtaining an estimation, the design of a learning model depends on one's limited knowledge about the dataset through manual observation and empirical trial-and-error analysis. After the model has been constructed with a configuration, it then learns by altering its parameters' values to fit itself to the distribution of the data. In this case of traditional machine learning and deep learning methods, with the more data provided, the closer the learned distribution is. In contrast, if the distribution of the temporal lengths of cell events is known, reference points can be used to determine appropriate hyper-parameters, particularly the sequence length, for each dataset. With that knowledge from unsupervised learning with the models M1 and M2, we can build model M3 that only requires 18 frames for training, instead of 110 frames, to achieve a much higher F1-score than published supervised methods (Table III). A model M3, with a sequence length of 8 and a training set of 18 frames, achieved F1-scores of 0.908-0.917, which is about 14% higher than the F1-scores of 0.795-0.800 of a supervised ConvLSTM model [29]. This is a huge saving in terms of manual effort for annotating cell imaging data. Although an unsupervised method has been developed [29], its performance was far lower than that of our method. In addition, giving the learning process a direction to guide the changes of the parameters is always desirable as we can control what the model's behaviors.

Supervised learning is affected by the ratio of positive and negative instances in the training data. In future work, we will improve our approach with a new method to balance the target cell events, events similar to the target cell events, and those that are not cell events. This can be done by analyzing the predicted cell events in the output of M2, $e'_t$. We will also optimize the loss function such that it accounts for the imbalance between positive and negative instances in our problem with limited annotations. Other methods to improve generalization and reduce overfitting, such as batch normalization [52], will also be adapted and incorporated into building the supervised model M3.

# VI/ Conclusion

We propose an automated unsupervised deep learning method to analyze cell videos to determine an optimal temporal length for the preparation of training data and model designing of the subsequent supervised learning for cell event detection. Our results show that it is possible to approximate the distribution of temporal lengths of cell events optimized to a video and train a model for mitosis detection with much fewer annotated frames than the traditional supervised methods. We suggest our method can be applied in biological experiments to minimize human labor for manual annotation.